\documentclass[letterpaper]{article}
\usepackage[preprint]{aaai2027}
\usepackage[hyphens]{url}
\usepackage{graphicx}
\usepackage{caption}
\usepackage{natbib}
\usepackage{booktabs}
\usepackage{amsmath}
\usepackage{amssymb}
\newcommand{\method}{\textsc{CASD}}

\title{\method{}: Chunk-Aligned Semantic Distillation for Multi-Stage Robot Manipulation}
\author{
Tinghe Ding\equalcontrib,
Jiahao Li\equalcontrib,
He Wang\corresponding
}
\affiliations{
Ant Group, China\\
\{tinghe.dth, ljh488565, he.wang\}@antgroup.com
}

\begin{document}

\maketitle
\pagestyle{plain}
\thispagestyle{plain}

\begin{abstract}
An action chunk can span several stages of a manipulation task, yet a label
for its first step describes only the current stage. We introduce Chunk-Aligned
Semantic Distillation (\method{}), which derives semantic targets
for entire action chunks. An offline
vision--language model segments demonstrations into described stages. Their
occupancy within each action chunk determines a weighted semantic target,
including transitions between stages. A CASD generator learns to predict this
target from the current observation, robot state, and task instruction. We
then freeze the generator and train a policy conditioned on its predictions.
The semantic branch runs once per policy query, without online VLM calls or
reasoning-trace decoding. Teacher matching on annotated LIBERO training
episodes is above chance for both single-stage and boundary-crossing chunks.
We evaluate three Fast-WAM variants and a DreamZero integration across
four benchmarks, including distribution shifts on LIBERO-Plus. Compared with
published references, IDM+\method{} reaches 98.9\% versus 98.0\% average
success on LIBERO, while Uncond falls below its reference. Joint+\method{}
reaches 93.0\% versus 90.6\% on RoboTwin 2.0, and DreamZero+\method{} reaches
a 47.9\% four-category MolmoSpaces manipulation average versus 40.7\%.
Performance varies across backbone integrations.
\end{abstract}

\section{Introduction}

Many vision--language--action (VLA) policies predict a chunk of future actions
at each query \citep{zhao2025cotvla,huang2026fastthinkact}. World Action Models
(WAMs) also model future visual states
\citep{DBLP:journals/corr/abs-2603-16666,DBLP:journals/corr/abs-2602-15922}.
Chunking reduces the number of policy queries. In multi-stage tasks,
however, the immediate objective can change within a predicted chunk while
the overall instruction stays fixed. A label for the current step may then
omit stages that the chunk will reach.

Consider placing a bowl in a drawer and then closing it. A chunk near the
placement boundary may contain transporting the bowl, releasing it, and
reaching toward the drawer handle. Labeling that chunk only as \emph{transport}
omits the upcoming release; labeling it only as \emph{close} skips an unfinished
interaction. Even a list of the overlapping stages leaves out how much of the
chunk each stage occupies. This motivates a semantic target defined over the
same time interval as the actions it supervises.

Language-conditioned action hierarchies \citep{belkhale2024rth} and progress
models \citep{chen2025sarm,yan2026progressvla,feng2026procvlm} provide temporal
structure for control. Visual chain-of-thought (CoT) and latent planning use
intermediate goals or plans to guide actions
\citep{zhao2025cotvla,huang2026fastthinkact}, while privileged distillation
transfers demonstrated future information to a current-only policy
\citep{fang2026pfd}. We focus on supervision at the policy's prediction
timescale, describing both the stages that overlap an action chunk and
the fraction of the chunk occupied by each.

A segmented demonstration supplies the required supervision. For a query at
time $t$, the fraction of its valid future action chunk spent in each stage
defines a chunk-level context: a single stage away from a boundary and a
weighted mixture near one. The weights depend on the demonstrated future, so
they are privileged information unavailable during deployment.

We transfer this context to the policy through Chunk-Aligned Semantic
Distillation (\method{}). An offline VLM-plus-action pipeline divides
training trajectories into contiguous stages with free-form descriptions.
We encode the descriptions and mix their semantic representations according
to their chunk occupancies. A CASD generator predicts this target from the
current observation, robot state, and task instruction. The target specifies
within-horizon stage composition, using free-form descriptions without a
global vocabulary of stage classes.

Stage A trains the CASD generator against this fixed teacher, separately from
the policy. Stage B freezes it and trains the video--action policy on its
predictions. Cross-attention adapters supply these tokens to the backbone.
The policy consequently receives predicted semantic conditions during both
training and deployment. The generator predicts all semantic slots in parallel
once per policy query; its branch makes no online VLM calls and generates no
textual reasoning trace or subgoal image.

We evaluate whether the generator recovers the teacher context and whether
that context helps execution. Teacher matching on annotated LIBERO training episodes
is above chance for both single-stage and boundary-crossing chunks. Policy
evaluation covers three Fast-WAM variants, distribution shifts on LIBERO-Plus,
bimanual tasks on RoboTwin, and MolmoSpaces with DreamZero.
The LIBERO average improves for IDM and Joint but declines for Uncond.

Our contributions are:
\begin{itemize}
    \item Chunk-level semantic supervision that weights stage descriptions by
    their occupancy in the action horizon, including chunks that cross stage
    boundaries.
    \item A current-input predictor and policy-conditioning interface that
    transfer this supervision to control, with the same frozen predictor
    supplying semantic conditions during policy learning and deployment.
    \item Evaluations of teacher matching, three Fast-WAM variants,
    distribution shifts on LIBERO-Plus, bimanual tasks on RoboTwin, and a
    DreamZero integration, including settings where performance declines.
\end{itemize}

\section{Related Work}

\paragraph{Stages, subgoals, and progress.}
Hierarchical control uses grounded skills \citep{ahn2022saycan}, executable
programs \citep{liang2022code}, and language/video plans \citep{ajay2023hip}.
RT-H predicts fine-grained language motions from observations and the task
instruction, then conditions action prediction on these motions within the
same imitation policy \citep{belkhale2024rth}. PALO searches over language
decompositions and ordered trajectory partitions, using a pretrained policy's
action-reconstruction error to choose conditions that fit the target
demonstrations \citep{myers2024palo}. PERIA combines language sub-instructions
with imagined visual subgoals \citep{ni2024peria}. UVD discovers visual subgoals
by detecting changes in pretrained visual embeddings along demonstrations
\citep{zhang2023uvd}, and Long-VLA selects visual context using motion and
interaction phases \citep{fan2025longvla}. SARM, ProgressVLA, and ProcVLM
study stage-aware rewards or progress supervision
\citep{chen2025sarm,yan2026progressvla,feng2026procvlm}.
\method{} weights free-form stage descriptions by their occupancy in the
action chunk. The target describes the composition of the prediction horizon:
queries with the same current-stage label can receive different targets,
and changing the horizon can change the target at the same observation.
This places semantic supervision at the timescale of action prediction,
without requiring a global stage vocabulary.
The distinction is the supervision target: stage composition within the action
horizon, rather than a scalar measure of episode completion.

\paragraph{Plans and reasoning for control.}
CoT-VLA autoregressively predicts future images as visual subgoals
\citep{zhao2025cotvla}, while ThinkAct learns visual latent plans
\citep{huang2025thinkact}. Fast-ThinkAct distills linguistic reasoning and
visual trajectory information into compact latents, with spatial tokens that
predict trajectory waypoints in parallel \citep{huang2026fastthinkact}.
ACoT-VLA uses action-space reference trajectories to guide action prediction
\citep{zhong2026acotvla}. ERVLA uses reasoning dropout to learn from embodied
CoT without decoding it during action prediction \citep{sun2026ervla}; ZR-0
likewise trains an action expert that can bypass reasoning-trace generation
\citep{li2026zr0}. \method{} directly predicts semantic context once per query,
without online VLM calls, text decoding, or subgoal-image generation in this
branch. Its teacher specifies which interactions overlap the action chunk and
their relative durations. It omits explicit paths and stage order; the policy
also receives current observations, robot state, and the task instruction.
Identical descriptions and occupancies yield the same teacher even when
execution order differs.

\paragraph{Privileged distillation.}
PFD transfers future information through the difference between
future-conditioned and current-only action-denoising outputs, fitting this
residual with a student adapter \citep{fang2026pfd}. \method{} constructs a
fixed teacher from demonstrated stage descriptions and occupancies,
independently of policy-teacher action outputs. The generator predicts this
target from current inputs and is then frozen for policy training.
Fast-ThinkAct also freezes its learned planner \citep{huang2026fastthinkact};
freezing and parallel token prediction are shared design choices. The
supervision differs: \method{} distills stage-duration mixtures rather than
reasoning traces or trajectories. Training the policy on generator predictions
exposes it to the same conditioning source, including prediction errors,
that it receives at deployment.
Demonstrated future boundaries are used only to construct training targets;
they are not inputs to the deployed generator.

\paragraph{World models for action prediction.}
OpenVLA learns observation-to-action mappings through robot pretraining
\citep{kim2025openvla}. WAMs additionally learn visual dynamics: Motus and
LingBot-VA couple video and action modeling \citep{bi2025motus,lingbot-va2026},
and DreamZero adapts a pretrained video-diffusion backbone for control
\citep{DBLP:journals/corr/abs-2602-15922}. Fast-WAM separates video co-training
from future-video generation at inference
\citep{DBLP:journals/corr/abs-2603-16666}. Its Joint and IDM variants denoise
video and actions together or sequentially, respectively, while Uncond uses
current-frame features. RIFT constructs future-conditioning
features in one pass \citep{zhang2026rift}. \method{} adds semantic conditioning
while retaining the backbone objectives and generation procedure. In Fast-WAM,
both experts receive the predicted tokens through separate projections and
adapters. The generator runs once per query, and adapters run at each
denoising step.
The action policy retains responsibility for selecting the executed sequence.
\section{Method}

\subsection{Chunk-Aligned Supervision}

\begin{figure*}[!t]
    \centering
    \includegraphics[width=\textwidth]{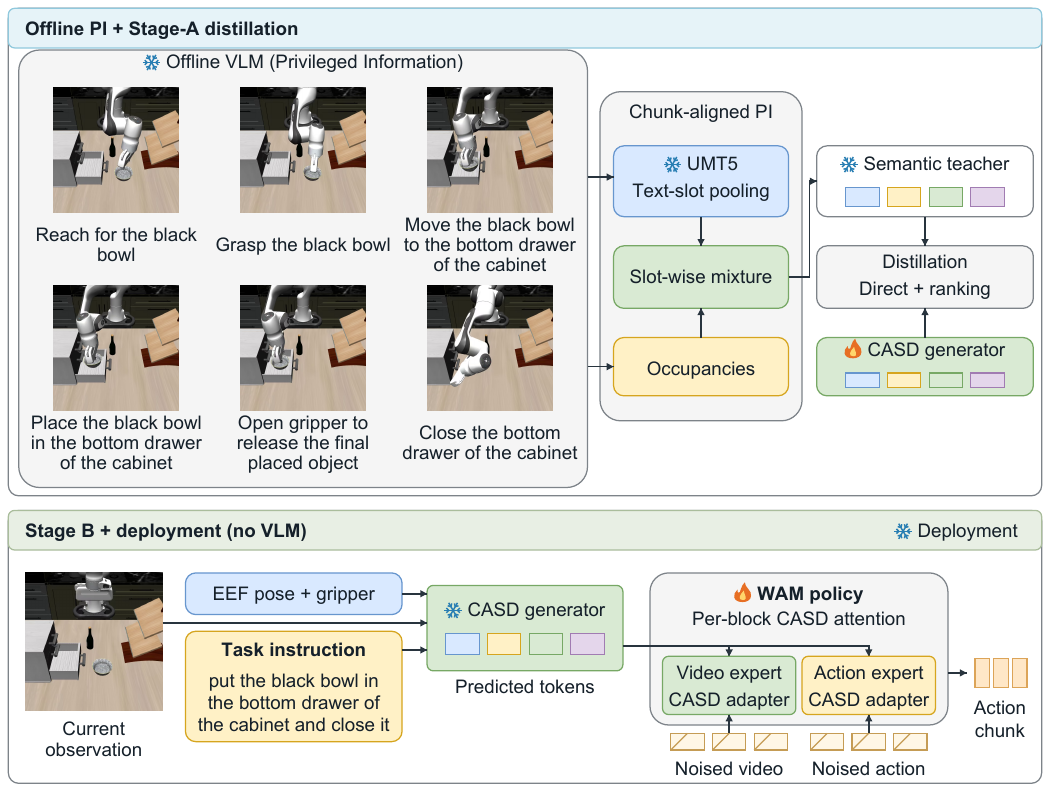}
    \caption{Two-stage privileged semantic distillation in \method. Offline
    stage annotations and chunk occupancies define a fixed semantic teacher.
    Stage A trains the CASD generator to predict these tokens from current
    inputs. Stage B freezes the generator and conditions the video--action
    policy on its predictions. The offline annotations and future-dependent
    targets are absent at deployment.}
    \label{fig:phase-framework}
\end{figure*}

We consider demonstrations pairing a task instruction $\ell$ with a trajectory
$\{(o_t,s_t,a_t)\}_{t=0}^{T-1}$ of observations, robot states, and actions. The
policy predicts an action chunk from $x_t=(o_t,s_t,\ell)$. We first define its
semantic supervision independently of the backbone, then describe prediction
and policy conditioning. Figure~\ref{fig:phase-framework} summarizes the two
training stages.

During training, we use GPT-5.4 as an offline VLM annotator
\citep{openai2026gpt54}. Given the task instruction, ordered frames, extracted
task entities, and action-derived boundary hints, it divides the episode into
an ordered, non-overlapping sequence of action-centered stages. Deterministic
postprocessing enforces full coverage and refines gripper transitions. We write
the semantic annotation as $\{([b_k,e_k),u_k)\}_{k=1}^{K}$, where $u_k$ is a
free-form stage description. Each policy query at time $t$ is paired with an
$H$-step demonstration action chunk. Let
$\mathcal{C}_t$ contain its valid, non-padded steps, and let $g_\tau$ be the
annotated stage at step $\tau$. We define the chunk occupancy of stage $k$ as
\begin{equation}
 w_{t,k}=
 \frac{|\{\tau\in\mathcal{C}_t:g_\tau=k\}|}
 {|\mathcal{C}_t|},
 \qquad \sum_k w_{t,k}=1.
 \label{eq:privileged-phase}
\end{equation}
For Fast-WAM, we use $H=32$ and include every annotated stage that overlaps the valid chunk.
The target represents all stages in the chunk. For example, a 32-step chunk
with 24 transport actions and eight release actions has occupancies
$3/4$ and $1/4$. A chunk fully within transport has occupancy one for that
stage. Near the episode end, the denominator counts only the remaining valid
actions. These occupancies use the demonstrated future and serve as privileged
training labels.

\subsection{Chunk-Aligned Semantic Teacher}

Let the frozen UMT5 encoder \citep{DBLP:conf/iclr/ChungGRTFNC23} produce the
valid token sequence $E(u_k)=\{h_{k,j}\}_{j=1}^{L_k}$ for stage description
$u_k$. We represent each description with $N$ ordered slots by partitioning
the sequence into contiguous, non-empty intervals
$\{\mathcal{I}_{k,n}\}_{n=1}^{N}$ and mean-pool each interval:
\begin{equation}
 \bar h_{k,n}=
 \frac{1}{|\mathcal{I}_{k,n}|}
 \sum_{j\in\mathcal{I}_{k,n}}h_{k,j}.
 \label{eq:phase-slot-pooling}
\end{equation}
The slots follow relative positions in the text sequence, independently of the
number of stages in the chunk. Let $\mu\in\mathbb{R}^{D_e}$ be the mean of
pooled training-description vectors and
$R\in\mathbb{R}^{D_p\times D_e}$ a fixed dimension-reduction map. The
occupancy-weighted teacher for slot $n$ is
\begin{equation}
 T_{t,n}
 =R\left(\sum_k w_{t,k}\bar h_{k,n}-\mu\right),
 \qquad T_t\in\mathbb{R}^{N\times D_p}.
 \label{eq:fused-teacher}
\end{equation}
Because the occupancies sum to one, this equals the weighted mixture of the
per-stage projected tokens. Each target has the same size regardless of how
many stages overlap the chunk. The mixture records stage proportions but
does not encode their order. The policy also receives the current observation
and instruction.
The teacher is fixed before student learning. Token count and projection
settings are specified in the experimental setup, with full construction
details in the supplement.

\subsection{CASD Generator from Current Inputs}

The CASD generator maps the current visual latent $z_t$, robot state $s_t$, and
task instruction to $P_t$. A frozen VAE encodes the current frame, and an
observation MLP fuses its features with the robot state. This feature and
the projected mean-pooled instruction form a conditioning vector $c_t$.
The model produces $N$ parallel queries from $c_t$ and attends to the full
instruction sequence $H^\ell=E(\ell)$:
\begin{equation}
\begin{aligned}
 Q_t&=\operatorname{reshape}(W_q c_t),\\
 K^\ell&=V^\ell=W_{kv}H^\ell,\\
 A_t&=\operatorname{MHA}(Q_t,K^\ell,V^\ell;m),\\
 P_t&=F_{\mathrm{tok}}(Q_t+A_t).
\end{aligned}
 \label{eq:phase-student}
\end{equation}
Here $m$ masks padded instruction tokens, and $F_{\mathrm{tok}}$ is a layer
normalization followed by a linear projection.
Adding $Q_t$ to $A_t$ lets visual and proprioceptive features influence the
output directly, rather than only through text-attention weights. The $N$
queries are processed in parallel, and each corresponds to an ordered semantic
slot in Equation~\eqref{eq:phase-slot-pooling}. The generator predicts the
mixed target directly.

\subsection{Stage-A Semantic Distillation}

Stage A trains only the CASD generator, sampling frame anchors from annotated
training demonstrations. The VAE, UMT5 encoder, teacher transformation, and
policy remain frozen. Precomputed current-frame features use the same VAE and
preprocessing as the online path.

Let $\sigma_T$ denote the RMS scale stored with the teacher. Because teacher and
prediction slots have a fixed correspondence, we use normalized direct
regression:
\begin{equation}
 d(P,T)=\frac{\|P-T\|_F^2}{ND_p\sigma_T^2},
 \qquad
 \mathcal{L}_{\mathrm{dir}}
 =\frac{\sum_i\beta_i d(P_i,T_i)}
 {\sum_i\beta_i}.
 \label{eq:phase-direct-loss}
\end{equation}
Here $\beta_i\in\{0,1\}$ indicates whether sample $i$ has a non-padded
action step and passes the episode-length and stage-boundary checks.
All retained samples have unit weight; stored annotation confidence and
validation flags do not affect this mask.

To separate contexts within the same task, we consider other valid teacher
candidates $T_j$ from that task in the global minibatch, with
$d(T_i,T_j)\geq\delta$ and choose the one that maximizes the ranking violation
$d(P_i,T_i)-d(P_i,T_j)+\rho d(T_i,T_j)$. Let $T_i^-$ denote this candidate and
$\mathcal{I}^{-}$ the samples for which one exists. With
$d_i^+=d(P_i,T_i)$, $d_i^-=d(P_i,T_i^-)$, and
$\Delta_i=d(T_i,T_i^-)$, the ranking objective is
\begin{equation}
 \mathcal{L}_{\mathrm{rank}}
 =\frac{\sum_{i\in\mathcal{I}^{-}}\beta_i
 \left[d_i^+-d_i^-+\rho\Delta_i\right]_+}
 {\sum_{i\in\mathcal{I}^{-}}\beta_i}.
 \label{eq:phase-ranking-loss}
\end{equation}
The Stage-A objective is
$\mathcal{L}_A=\mathcal{L}_{\mathrm{dir}}
+\lambda_{\mathrm{rank}}\mathcal{L}_{\mathrm{rank}}$.
We use $\lambda_{\mathrm{rank}}=1$, $\delta=0.05$, and $\rho=0.5$.
Each loss is zero when its denominator is zero.

\subsection{Stage-B Policy Learning and CASD Injection}

Stage B loads the Stage-A checkpoint, freezes the CASD generator, and conditions
the policy on its predictions. The same token source is used during policy
training and deployment. Teacher targets and alignment losses are not used in
Stage B; the objective remains the original sum of visual-dynamics and action
flow-matching losses,
\begin{equation}
 \mathcal{L}_B=\mathcal{L}_{\mathrm{dyn}}+\mathcal{L}_{\mathrm{act}}.
 \label{eq:policy-stage-loss}
\end{equation}

In the Fast-WAM implementation \citep{DBLP:journals/corr/abs-2603-16666},
the video and action experts share predicted tokens $P_t$ and project them
separately to $C_{t,e}=g_e(P_t)\in\mathbb{R}^{N\times D_c}$. Each $g_e$ is an
independently learned affine map followed by GELU. Each block receives a
separately parameterized low-rank cross-attention adapter:
\begin{equation}
 \widetilde X_e^{(l)}
 =X_{e,\mathrm{std}}^{(l)}
 +\mathcal{A}_{p,e}^{(l)}(X_e^{(l)},C_{t,e}),
 \qquad e\in\{v,a\}.
 \label{eq:expert-routing}
\end{equation}
Policy parameters and adapters are optimized under the backbone losses while
the CASD generator stays frozen. We use $D_c=1024$ and rank-192 adapter
bottlenecks. The final adapter projection is initialized to zero, preserving
the pretrained function at initialization. Full architecture dimensions are
provided in the supplementary material.

\begin{figure*}[!t]
    \centering
    \includegraphics[width=1.0\textwidth]{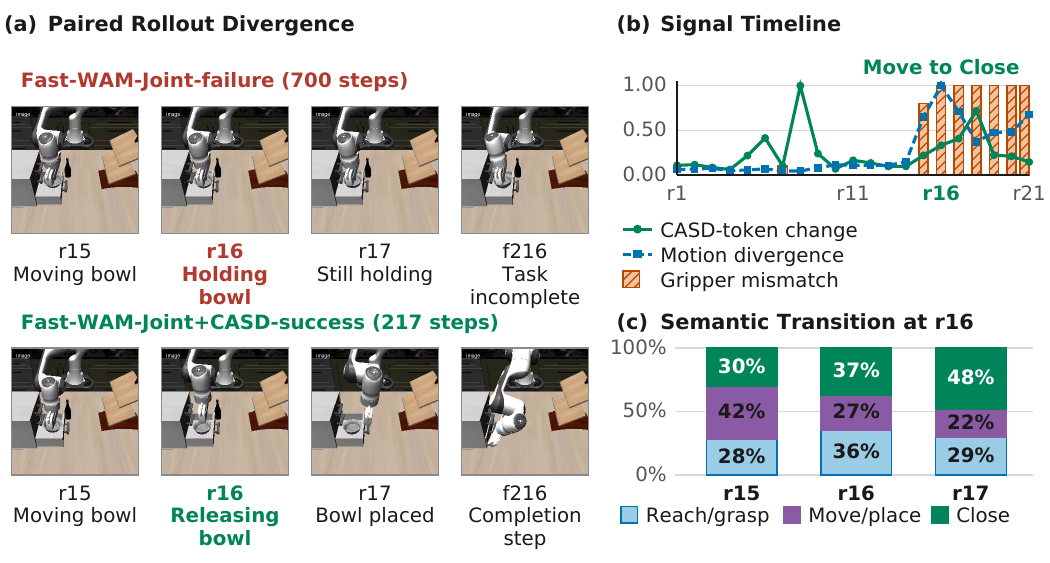}
    \caption{A paired LIBERO-Long rollout showing where the baseline and
    \method{} trajectories diverge. The Fast-WAM-Joint baseline times out,
    whereas Fast-WAM-Joint+\method{} succeeds from the same recorded initial
    observation with matched inference seeds. (a) By replan r16, \method{}
    has begun releasing the bowl while the baseline keeps holding it. (b) This behavioral
    difference coincides with the largest motion divergence and complete
    gripper-command disagreement; token and motion signals are normalized
    separately. (c) A post-hoc convex fit to task-specific teacher prototypes
    shows a shift from Move/place toward Close. The displayed weights describe
    the fit and do not enter the policy.}
    \label{fig:libero-paired-case}
\end{figure*}

\subsection{Inference}

Before each diffusion loop, the CASD generator computes the context once from
the latest visual latent, current robot state, and task instruction. The policy
reuses this context across diffusion steps and recomputes it after the
environment advances. The original backbone determines how video and action
are denoised: Joint generates them together, IDM generates video before
actions, and Uncond uses current-frame features. Across variants, CASD predicts all
semantic slots in parallel once per policy query, without online VLM calls,
text decoding, or subgoal-image generation. CASD adapters run at each
denoising step.

\section{Experiments}

\subsection{Experimental Setup}

\paragraph{Benchmarks.}
We evaluate on four benchmarks. LIBERO contains Spatial, Object,
Goal, and Long suites \citep{DBLP:conf/nips/LiuZGFLZS23}; LIBERO-Plus applies
seven controlled shifts spanning layout, camera, robot initialization,
language, lighting, background, and sensor noise
\citep{DBLP:journals/corr/abs-2510-13626}. RoboTwin 2.0 evaluates 50 bimanual
tasks under clean and domain-randomized conditions \citep{chen2025robotwin}.
MolmoSpaces-Bench spans manipulation and navigation in diverse indoor scenes
\citep{DBLP:journals/corr/abs-2602-11337}. We evaluate the four-category
Franka/DROID manipulation subset used by the leaderboard snapshot: Pick,
Pick-and-Place, Open, and Close \citep{molmospaces2026leaderboard}.
Policy tables report success rates (\%); Table~\ref{tab:phase-context-retrieval}
reports teacher-matching recall on a $[0,1]$ scale.

\paragraph{Model setup.}
On LIBERO and LIBERO-Plus, we evaluate \method{} with the Uncond, Joint, and
IDM Fast-WAM variants. RoboTwin 2.0 includes the same three \method{} variants,
alongside the three official Fast-WAM baselines.
These models build on Wan2.2-TI2V-5B
\citep{DBLP:journals/corr/abs-2603-16666,wanai2025wan22ti2v5b}.
The \method{} runs use a frozen 10,000-step Stage-A CASD generator and a fixed
seed-0 random-projection teacher with $N=4$ tokens of dimension $D_p=128$, with
\method{} routed to both video and action experts.

For cross-backbone evaluation, we integrate \method{} into
DreamZero (Wan2.1-I2V-14B) \citep{DBLP:journals/corr/abs-2602-15922}; both its
Franka pretraining and our \method{}-augmented training use DROID
\citep{DBLP:conf/rss/KhazatskyP0BDKN24}.
Our \method{} policies use predicted semantic conditions in Stage B,
matching the context source at evaluation.
Tables~\ref{tab:libero-main} and~\ref{tab:robotwin} compare our runs with
published results for the corresponding backbone variants and benchmark
settings. The LIBERO-Plus references are separately evaluated checkpoints
reported by RIFT. These comparisons do not control for differences in
training runs. We express differences in percentage points (pp). The
supplement provides the LIBERO pipeline, metric definitions, and result sources.

\begin{table}[!ht]
\centering
\small
\setlength{\tabcolsep}{3pt}
\begin{tabular}{@{}lrrrr@{}}
\toprule
Subset & Queries & Share & Chance Recall &
CASD Recall \\
\midrule
Pure & 5,312 & 55.87\% & 0.185 & 0.430 \\
Mixed & 4,195 & 44.13\% & 0.174 & 0.765 \\
\quad Mixed-2 & 3,643 & 38.32\% & 0.175 & 0.761 \\
\quad Mixed-3+ & 552 & 5.81\% & 0.166 & 0.790 \\
\bottomrule
\end{tabular}
\caption{Episode-local teacher matching at annotated training phase starts.
Pure queries cover one stage in the valid part of the following 32-step chunk;
Mixed queries cross stage boundaries. Both recalls are query averages;
Mixed-2 and Mixed-3+ partition Mixed.}
\label{tab:phase-context-retrieval}
\end{table}
\setcounter{table}{2}
\begin{table*}[!t]
\centering
\small
\setlength{\tabcolsep}{2.4pt}
\begin{tabular}{lrrrrrrrr}
\toprule
Method & Camera & Robot Init. & Language & Light & Background & Noise & Layout & Pooled \\
\midrule
NORA & 2.2 & 37.0 & 65.1 & 45.7 & 58.6 & 12.8 & 62.1 & 39.0 \\
WorldVLA & 0.1 & 27.9 & 41.6 & 43.7 & 17.1 & 10.9 & 38.0 & 25.0 \\
UniVLA & 1.8 & 46.2 & 69.6 & 69.0 & 81.0 & 21.2 & 31.9 & 42.9 \\
$\pi_0$ & 13.8 & 6.0 & 58.8 & 85.0 & 81.4 & \textbf{79.0} & 68.9 & 53.6 \\
$\pi_0$-FAST & \textbf{65.1} & 21.6 & 61.0 & 73.2 & 73.2 & 74.4 & 68.8 & 61.6 \\
RIPT-VLA & 55.2 & 31.2 & 77.6 & 88.4 & \textbf{91.6} & 73.5 & 74.2 & 68.4 \\
OpenVLA-OFT-M & 55.6 & 21.7 & 81.0 & 92.7 & 91.0 & 78.6 & 68.7 & 67.9 \\
\midrule
Fast-WAM$^{\dagger}$ & 16.5 & 43.4 & 67.8 & 80.1 & 51.6 & 37.9 & 61.2 & 49.7 \\
Fast-WAM+\method{} & 28.8 & 49.9 & 75.2 & 83.8 & 54.5 & 51.0 & 64.8 & 57.2 \\
\addlinespace[2pt]
Fast-WAM-Joint$^{\dagger}$ & 38.7 & 63.9 & 93.1 & \textbf{94.7} & 55.7 & 57.7 & 77.5 & 68.1 \\
Fast-WAM-Joint+\method{} & 46.0 & 70.7 & \textbf{94.5} & 94.5 & 62.3 & 60.3 & 81.3 & 72.2 \\
\addlinespace[2pt]
Fast-WAM-IDM$^{\dagger}$ & 46.2 & 69.1 & 94.2 & 92.8 & 56.4 & 61.8 & \textbf{81.5} & 71.4 \\
Fast-WAM-IDM+\method{} & 50.3 & \textbf{77.0} & 90.6 & 92.8 & 67.1 & 62.8 & 81.2 & \textbf{73.9} \\
\bottomrule
\end{tabular}
\caption{LIBERO-Plus success scores (\%). Pooled is the variant-micro rate over
10,030 benchmark variants; external comparison rows follow the benchmark report
\citep{DBLP:journals/corr/abs-2510-13626}. $^{\dagger}$~RIFT-reported
Fast-WAM references \citep{zhang2026rift}; \method{} rows are ours.}
\label{tab:libero-plus}
\end{table*}
\setcounter{table}{1}

\subsection{Teacher Matching from Current Inputs}

Using only current inputs, we evaluate the frozen CASD generator on
annotated LIBERO training episodes. After prediction, each query is matched against
teacher targets at annotated phase starts in the same episode, ranked by
negative normalized MSE. Recall@1 assigns credit $1/m$ when the correct
target is among $m$ candidates tied for the best score, and zero otherwise. Chance Recall averages $1/N_e$ over queries,
where $N_e$ is the episode gallery size.

The evaluation contains 9,507 phase-start queries from 1,712 episodes
(5.55 per episode). Queries are grouped by the stages in their valid following
32-step chunk, as defined in Table~\ref{tab:phase-context-retrieval}.
Recall is 2.3$\times$ chance for Pure queries (0.430 vs.\ 0.185) and
4.8$\times$ for Mixed-3+ (0.790 vs.\ 0.166). These training-set diagnostics
show recovery of boundary-crossing contexts from current inputs; comparisons
between subsets also depend on the distinctiveness of their gallery targets.

\subsection{Downstream Policy Evaluation}

\paragraph{In-distribution performance on LIBERO.}
Table~\ref{tab:libero-main} shows average differences of $-3.2$, $+0.3$, and
$+0.9$ pp for Uncond, Joint, and IDM, respectively, using unrounded run
averages. IDM+\method{} reaches the highest average in the table (98.9\%).
Joint improves from 98.5\% to 98.8\%, with its largest gain on Long (+1.2 pp),
followed by Goal (+0.4 pp); Spatial and Object change by $-0.2$ and $-0.4$ pp.
The Uncond decline is largest on Goal ($-4.4$ pp). Two of the three CASD
variants score above their published references on average, with gains and
regressions across individual suites.

\paragraph{Illustrative paired case.}
Figure~\ref{fig:libero-paired-case} examines trial 2 of ``put the black bowl in
the bottom drawer of the cabinet and close it.'' The Fast-WAM-Joint baseline
succeeds in 49/50 trials and \method{} in 50/50; trial 2 is their paired
failure/success case, with the same recorded initial observation and matched
seeds at all 22 common replans.

By replan r16, \method{} has begun releasing the bowl while the baseline
keeps holding it. The r15$\rightarrow$r16 transition has token change
0.060, non-gripper action RMSE 0.460, and disagreement on all 10 aligned gripper
commands. In the visualization-only fit, Move/place falls from 42\% to 27\%
and then 22\%, while Close rises from 30\% to 37\% and 48\%. \method{} begins
releasing at policy step 152 and completes the task at zero-indexed step 216
(217 executed steps); the baseline releases 84 steps later and times out at 700
steps. This semantic transition coincides with the largest motion
divergence among the 22 common replans.

\begin{table}[!htb]
\centering
\small
\setlength{\tabcolsep}{1.6pt}
\begin{tabular}{@{}l@{\hspace{1.2pt}}cccccc@{}}
\toprule
Method & Emb. PT & Spatial & Object & Goal & Long & Avg. \\
\midrule
$\pi_{0.5}$ & $\checkmark$ & 98.8 & 98.2 & 98.0 & 92.4 & 96.9 \\
GR00T-N1.7 & $\checkmark$ & 97.7 & 98.5 & 97.5 & 94.4 & 97.0 \\
Motus & $\checkmark$ & 96.8 & 99.8 & 96.6 & 97.6 & 97.7 \\
Qwen-RM-scratch & $\times$ & -- & -- & -- & -- & 98.2 \\
LingBot-VA & $\checkmark$ & 98.5 & 99.6 & 97.2 & \textbf{98.5} & 98.5 \\
\midrule
Fast-WAM$^{\dagger}$ & $\times$ & 98.2 & \textbf{100.0} & 97.0 & 95.2 & 97.6 \\
Fast-WAM+\method{} & $\times$ & 97.6 & 94.8 & 92.6 & 92.8 & 94.5 \\
\addlinespace[2pt]
Fast-WAM-Joint$^{\dagger}$ & $\times$ & \textbf{99.6} & 99.4 & 98.2 & 96.8 & 98.5 \\
Fast-WAM-Joint+\method{} & $\times$ & 99.4 & 99.0 & \textbf{98.6} & 98.0 & 98.8 \\
\addlinespace[2pt]
Fast-WAM-IDM$^{\dagger}$ & $\times$ & 98.8 & 97.8 & 97.8 & 97.6 & 98.0 \\
Fast-WAM-IDM+\method{} & $\times$ & \textbf{99.6} & 99.8 & 98.2 & 97.8 & \textbf{98.9} \\
\bottomrule
\end{tabular}
\caption{LIBERO success rates (\%). Emb. PT: embodied pretraining; Qwen-RM:
Qwen-RobotManip; unsuffixed Fast-WAM: Uncond. Bold marks the column best.
$\dagger$ marks official Fast-WAM results \citep{DBLP:journals/corr/abs-2603-16666}.
External rows follow $\pi_{0.5}$ \citep{intelligence2025pi05}, GR00T-N1.7
\citep{nvidia2026grootn17libero}, Motus \citep{bi2025motus}, Qwen-RM
\citep{qwenrobotmanip2026}, and LingBot-VA \citep{lingbot-va2026}.}
\label{tab:libero-main}
\end{table}
\setcounter{table}{3}

\paragraph{Robustness under LIBERO-Plus shifts.}
Table~\ref{tab:libero-plus} reports \method{}-augmented Fast-WAM results under
seven controlled visual, robot-state, and language shifts.

Our \method{} runs reach 57.2\%, 72.2\%, and 73.9\% pooled success for
Fast-WAM, Joint, and IDM, respectively. RIFT reports external references of
49.7\%, 68.1\%, and 71.4\%. Because the checkpoints and training recipes differ,
the $+7.5$/$+4.1$/$+2.5$ pp differences are cross-report comparisons. Our Joint model is
higher on six of seven shifts, with the largest gaps for camera viewpoint
(+7.3 pp), robot initial state (+6.8 pp), and background texture (+6.6 pp).
Camera viewpoint remains its weakest category (46.0\%), well below language
and lighting (both 94.5\%). IDM+\method{} has the highest pooled rate among
our variants, although its language score is lower than Joint+\method{}.

\begin{table}[!htb]
\centering
\small
\setlength{\tabcolsep}{2.5pt}
\begin{tabular}{@{}lcrrr@{}}
\toprule
Method & Embodied PT & Clean & Rand. & Average \\
\midrule
$\pi_{0.5}$ & $\checkmark$ & 82.7 & 76.8 & 79.8 \\
Motus & $\checkmark$ & 88.7 & 87.0 & 87.8 \\
Qwen-RM-scratch & $\times$ & 88.7 & 88.4 & 88.6 \\
LingBot-VA & $\checkmark$ & \textbf{92.9} & 91.6 & 92.2 \\
\midrule
Fast-WAM$^{\dagger}$ & $\times$ & 91.9 & 91.8 & 91.8 \\
Fast-WAM+\method{} & $\times$ & 90.5 & 91.4 & 90.9 \\
Fast-WAM-Joint$^{\dagger}$ & $\times$ & 90.8 & 90.3 & 90.6 \\
Fast-WAM-Joint+\method{} & $\times$ & 92.7 & \textbf{93.3} & \textbf{93.0} \\
Fast-WAM-IDM$^{\dagger}$ & $\times$ & 91.2 & 91.3 & 91.3 \\
Fast-WAM-IDM+\method{} & $\times$ & 92.6 & 92.2 & 92.4 \\
\bottomrule
\end{tabular}
\caption{RoboTwin 2.0 success rates (\%) across 50 tasks. Clean and Rand. denote
Easy and Hard; Embodied PT denotes embodied pretraining; Qwen-RM abbreviates
Qwen-RobotManip. Bold marks the column best. $\dagger$ marks official Fast-WAM
baselines from \citet{DBLP:journals/corr/abs-2603-16666}; other external sources
are cited in the text.}
\label{tab:robotwin}
\end{table}

\paragraph{Bimanual robustness on RoboTwin 2.0.}
We report mean success across 50 tasks under clean (Easy) and domain-randomized
(Hard) evaluation. External results follow their source papers
\citep{intelligence2025pi05,bi2025motus,qwenrobotmanip2026,lingbot-va2026}.

Joint+\method{} exceeds the official Joint baseline by 1.9 pp on Clean and
3.0 pp on Rand. (+2.4 pp average).
Uncond+\method{} reaches 90.9\% average (90.5\% Clean and 91.4\% Rand.).
IDM+\method{} scores 92.6\% Clean and 92.2\% Rand. (92.4\% average).
Joint+\method{} leads
Table~\ref{tab:robotwin} in Rand. (93.3\%) and average (93.0\%).

\begin{table}[!htb]
\centering
\small
\setlength{\tabcolsep}{3pt}
\begin{tabular}{@{}lrrrrr@{}}
\toprule
Method & Pick & P\&P & Open & Close & Avg. \\
\midrule
PRTS-Droid & 53.8 & 27.7 & 12.0 & \textbf{76.3} & 42.5 \\
MolmoAct2 DROID & 48.3 & 23.3 & 11.7 & 71.3 & 38.7 \\
$\pi_{0.5}$ DROID & 36.4 & 13.6 & 22.7 & 65.1 & 34.5 \\
$\pi_0$ DROID & 16.2 & 12.5 & 11.0 & 53.1 & 23.2 \\
LAP-VLA & 24.9 & 6.6 & 11.4 & 45.9 & 22.2 \\
$\pi_0$-FAST DROID & 22.0 & 10.9 & 11.1 & 38.6 & 20.7 \\
Psi-R2 & \textbf{73.1} & \textbf{41.5} & 9.0 & 62.2 & 46.5 \\
DreamZero & 52.1 & 25.6 & 24.9 & 60.3 & 40.7 \\
DreamZero+\method{} & 59.7 & 36.1 & \textbf{35.1} & 60.8 & \textbf{47.9} \\
\bottomrule
\end{tabular}
\caption{Success rates (\%) on four MolmoSpaces manipulation categories.
P\&P denotes Pick-and-Place; navigation is excluded. External values follow the
official leaderboard snapshot \citep{molmospaces2026leaderboard}; the benchmark
is described by \citet{DBLP:journals/corr/abs-2602-11337}. The average is
computed over the four listed categories. Bold marks the best value in each column.}
\label{tab:molmospaces}
\end{table}

\paragraph{Integration into DreamZero.}
We integrate \method{} into DreamZero and evaluate four MolmoSpaces
manipulation categories; navigation is outside this comparison.

DreamZero+\method{} reaches 47.9\% average success, compared with 40.7\% for
the leaderboard DreamZero reference (+7.2 pp). The largest category differences
are on Pick-and-Place (+10.5 pp) and Open
(+10.2 pp). DreamZero+\method{} has the highest average in
Table~\ref{tab:molmospaces}.

\subsection{Discussion}

\paragraph{Semantic context at the action horizon.}
Teacher matching shows that \method{} can recover chunk-level context on
annotated training episodes. The paired rollout in
Figure~\ref{fig:libero-paired-case} illustrates how a predicted semantic
transition can accompany a change in manipulation behavior.

\paragraph{Interaction with the policy backbone.}
Joint and IDM exceed their published LIBERO and RoboTwin references, whereas
Uncond declines. All three exceed the LIBERO-Plus references. Semantic context
may complement each backbone's learned representations differently. Matched
comparisons of current-stage labels, equal-weight and occupancy-weighted
mixtures, and expert routing could identify which conditions benefit each
variant; the present reference comparisons do not isolate these factors.

\paragraph{Richer temporal predictions.}
The current teacher retains stage proportions but discards stage order.
Constructing separate mixtures over successive parts of the action horizon
would preserve coarse order while keeping a bounded token budget. A second
direction is to predict several possible semantic futures using a short
observation history. This could distinguish repeated visits to similar states
and avoid averaging incompatible contexts when several continuations are
plausible. Varying both the prediction horizon and the replanning interval
would test how much temporal detail control needs.

\paragraph{Learning from execution.}
Automatic annotations can contain grounding and boundary errors, and policy
rollouts can depart from the demonstrations used for distillation. Controlled boundary
perturbations during distillation could test sensitivity to annotation timing.
Separately, annotating policy rollouts would allow semantic prediction errors
to be measured during execution. Execution feedback could then guide refinement of the teacher or
adaptation of the generator. These extensions should be evaluated jointly for
task success and query latency, including generator and adapter costs under
matched hardware and sampling settings.

\section{Conclusion}

\method{} aligns semantic supervision with the time span of an action chunk.
A fixed teacher combines stage descriptions according to their occupancy,
and a generator predicts the resulting context from current inputs. Freezing
the generator gives policy learning and deployment the same source of
semantic conditions. Evaluations with three Fast-WAM variants and
DreamZero demonstrate its use across video--action generation schemes, with
task success depending on the backbone and benchmark. The central
contribution is a way to connect demonstrated task structure to the temporal
resolution of action prediction.
{\small
\bibliography{phase_refs}
}

\clearpage
\appendix
\section*{Supplementary Material}

These appendices detail the LIBERO annotation and distillation pipeline,
teacher-matching protocol, policy metrics, per-task RoboTwin results, and
the paired rollout in Figure~\ref{fig:libero-paired-case}.

\setcounter{table}{0}
\renewcommand{\thetable}{S\arabic{table}}

\section{Offline Phase Annotation}

\subsection{Annotation Unit and Raw Episode Inputs}

Each annotation covers one complete LIBERO training demonstration and its
benchmark instruction \citep{DBLP:conf/nips/LiuZGFLZS23}. The process reads the
four local sources in Table~\ref{tab:label-inputs}. The VLM receives the task
text, a frame contact sheet, and action-derived event hints. Episode identifiers,
raw actions, and the future $H{=}32$ action occupancy remain local; occupancy is
computed after annotation for Stage-A supervision.

The output for an episode is a temporally ordered sequence
\[
  \mathcal{S}=\{([b_k,e_k),u_k,q_k,v_k)\}_{k=1}^{K},
\]
where $[b_k,e_k)$ is a half-open frame interval, $u_k$ is a free-form,
action-first description, $q_k\in[0,1]$ is a stored annotation score initialized
from the returned confidence or a postprocessing rule, and $v_k$ is an audit flag.
Intervals are normalized to be contiguous, non-overlapping, positive in
length, and to cover the episode. The scores and flags are retained for
auditing. They do not set training weights, and the scores are not calibrated
probabilities.

\subsection{VLM-Visible Request}

\paragraph{Contact sheet.}
The sampler begins with 12 uniform indices from frame 0 through frame $T-1$,
including both endpoints.  Selected chronological action-hint frames may
replace at most four interior cells.  This retains global visual coverage
while adding observations near likely transitions.  Every cell is numbered
left-to-right as \texttt{i/12}. The percentage
printed inside a cell is its ordinal position in the strip; after a hint
replacement it should therefore be read as an ordering cue, not an exact
source timestamp.  The accompanying text hint gives the exact source frame
and its true normalized percentage.

\paragraph{Text prompt.}
The same user message contains the complete selected task instruction;
objects, destinations, and extra action verbs extracted from that string; and
the chronological hint list.  A hint is rendered only as
\texttt{t=.., frame .., gripper opens/closes}, \texttt{velocity peak}, or
\texttt{stillness}.  Hint scores, raw action values, and detection thresholds
are omitted.  The prompt defines visually observable transition cues, asks
the boundary to be the first frame at which a gripper change, object lift-off,
arm-direction reversal, or destination entry becomes visible, and asks for the
fewest ordered stages needed to complete the task. Each action verb in the task instruction must be represented, with separate
Place and Release descriptions for every pick-and-place operation.  Descriptions should contain 5--12 words, start with
an action verb, and preserve task nouns.  Suggested phase counts are 2--4 for
a simple single-verb task, 4--6 for one pick-and-place target, and 5--8 for a
multi-verb task; these are guidance, not fixed classes.

The request uses one multimodal user message containing the prompt and
base64-encoded JPEG. The launcher requests the \texttt{gpt-5.4} model alias
\citep{openai2026gpt54}, temperature 0.1, and at most 800 completion tokens.
Sampling seed, top-$p$, and schema mode use the service defaults. The client
retries failed requests up to four times and adapts unsupported parameter names.

The requested response is:
\begin{quote}
\small\ttfamily\raggedright
\{"phase\_segments":[\{"subtask": string,\\
"start\_ratio": float, "end\_ratio": float,\\
"confidence": float\}]\}
\end{quote}
The VLM returns the ordered descriptions, coarse normalized boundaries, and
per-segment confidence. The deterministic postprocessor derives final frame
intervals, validity flags, and Release-event metadata.

\begin{table*}[t]
\centering
\small
\setlength{\tabcolsep}{4pt}
\begin{tabular}{@{}p{1.07in}p{1.27in}p{3.26in}p{0.77in}@{}}
\toprule
Local source & Fields read & Use before or after the VLM call & VLM-visible? \\
\midrule
\raggedright\texttt{episodes.jsonl} &
\raggedright\texttt{episode\_index}, \texttt{length}, \texttt{task} or
\texttt{tasks[0]} &
Selects the episode and its single authoritative instruction.  A supplied
episode list may override the selected task string. &
\raggedright Task only \tabularnewline
\addlinespace
\texttt{info.json} &
Video features and action names &
Selects the first available exterior/non-wrist video when possible and
recognizes the seven-dimensional LIBERO action layout. &
\raggedright No \tabularnewline
\addlinespace
Episode Parquet &
\raggedright\texttt{frame\_index}, \texttt{action} &
Rows are sorted by frame.  The first six action dimensions yield an
action-space motion magnitude; the seventh yields gripper events.  Raw
values, units, and thresholds stay local. &
\raggedright Derived hints only \tabularnewline
\addlinespace
Episode video &
Selected RGB frames &
Twelve frames are resized to 144 pixels in height, concatenated horizontally,
and JPEG-encoded at quality 92.  The full video is not uploaded. &
\raggedright Contact sheet only \tabularnewline
\bottomrule
\end{tabular}
\caption{Inputs used for automatic phase annotation. ``Derived hints'' contain
an event type and time but not the underlying action vector. Because the action
dimensions have no stated SI units, motion is reported in action space.}
\label{tab:label-inputs}
\end{table*}

\subsection{Action-Hint Extraction}

For the seven-dimensional LIBERO action, the motion signal is the Euclidean
norm of the first six coordinates.  A \emph{velocity peak} exceeds the
episode's 70th percentile and is strictly larger than the values one and two
frames on either side.  A \emph{stillness} candidate has a ten-entry local
mean below the 20th percentile; nearby candidates with gaps of at most eight
frames are clustered and represented by a cluster center.

The gripper is considered open above 0.5 and closed otherwise.  A transition
is emitted only after the new Boolean state persists for three consecutive
frames; its event location is the first frame of that confirmed run, and the
initial stable state is not an event.  For the default 12-frame request, at
most eight gripper transitions, five velocity peaks, and four stillness
centers are considered.  Velocity peaks within ten frames of an already
selected event and stillness centers within twelve frames are skipped.
Candidates are then prioritized using internal scores 1.0, 0.6, and 0.45,
respectively; candidates within six frames of a retained event are
deduplicated, and at most 16 hints remain. These scores affect local selection
only; they are separate from the returned confidence used to initialize $q_k$.

\subsection{Generic Response Postprocessing}

The parser accepts a bare JSON object, a fenced JSON object, or the first
balanced object followed by prose.  An unparseable response is logged as an
episode error; there is no semantic re-prompt in this LIBERO path.  A parsed
response is converted to intervals as follows:
\begin{enumerate}
  \item Clamp the start and end ratios to $[0,1]$. If an end does not exceed
        its start, replace it by $\min(1,\text{start ratio}+1/K)$.
  \item Set the first start to 0.  For every later phase, discard its proposed
        start as an independent boundary and set it to the previous final end.
        Before snapping, a non-final end is
        $\operatorname{round}(\texttt{end\_ratio}\,T)$; the final end is $T$.
  \item Interpret each boundary using the \emph{next} description.  Before
        Release/Open it may snap only to a gripper-open event; before
        Grasp/Close only to gripper-close; otherwise only to a velocity peak
        or stillness event.  The nearest compatible hint is used only within
        radius $\max(5,\lfloor T/30\rfloor)$.
  \item Reserve at least one frame for every remaining phase and prefer a
        three-frame minimum when feasible.  Renumber phases and enforce
        contiguous half-open coverage through frame $T$.
\end{enumerate}
\FloatBarrier
Descriptions are stripped and truncated at 120 characters. Missing confidence
defaults to 0.85. The field
\path{boundary_evidence=vlm_vision+action_snap} records the postprocessing path;
by itself, it does not mean that a boundary was snapped to a nearby hint.

\subsection{Deterministic Place--Release Reconstruction}

Release receives a second, action-only refinement because sparse images do not
reliably identify the opening instant.  Any adjacent VLM Place--Release pair
is first collapsed back to its combined placement window.  Place candidates
are recognized from \texttt{place}, \texttt{put}, \texttt{insert},
\texttt{set}, \texttt{drop}, \texttt{hang}, or \texttt{stack}.  The
postprocessor then re-detects the complete, uncapped sequence of stable
gripper-open events from the raw local action stream.

For a Place interval, a candidate event must be unused, occur no earlier than
its start minus $r$, and occur no later than its end plus $r$ or the next Place
start, whichever comes first, where $r=\max(5,\lfloor T/30\rfloor)$.  The last
eligible event in this window is chosen.  Place ends at that frame; Release
starts there and ends no later than the original placement end or the next
stable close.  It must include the three-frame open confirmation, borrowing
frames from the following interval when necessary. The reconstructed segment
is assigned a stored score of 1.0, a valid audit flag, and evidence of the form
\path{stable_gripper_close_to_open@F...:confirm=3}.

Each ordered Place must use a distinct later release event.  A task that
requires placement but has no Place description, lacks usable action data, or
cannot form a consistent interval is excluded and logged.  Two recoverable
cases---a missing Release for a detected Place and a coarse multi-object Place
description---retain the episode but set the stored scores of that Place and
all later segments to zero, with \texttt{phase\_valid=false}. If motion
continues after a final verified Release and the gripper closes again, the
remaining tail is likewise stored as a completion segment with score zero
and a false audit flag. Only Release stages matched to an action event are retained.

\subsection{Stored Annotation Record}

The stored episode includes the selected task, final segments, annotation
scores and audit flags, hint frames and types, generator metadata, and the
terminal-release audit. The stored episode score is the mean of its segment
scores. Fatal cases enter a separate error log, while recoverable fallback
cases remain in the annotation record.

Stage A uses a binary sample mask $\beta_i$. The loader requires contiguous
stage coverage and checks the recorded episode length against action-padding
metadata when available. A sample receives $\beta_i=1$ if these checks pass
and its action chunk has at least one non-padded step, and zero otherwise.
Stored scores and \texttt{phase\_valid} flags do not enter this mask.
Consequently, a zero-score or audit-flagged stage can still contribute to the
occupancy mixture and the training loss.

\begin{table}[htbp]
\centering
\small
\setlength{\tabcolsep}{2.5pt}
\begin{tabular}{lrrrrr}
\toprule
Suite & Episodes & Queries & Pure & Mixed & Invalid \\
\midrule
Spatial & 434 & 2,367 & 1,173 & 1,194 & 7 \\
Object  & 457 & 2,457 & 1,377 & 1,080 & 3 \\
Goal    & 433 & 1,928 & 1,088 &   840 & 1 \\
Long    & 388 & 2,755 & 1,674 & 1,081 & 115 \\
\midrule
Total   & 1,712 & 9,507 & 5,312 & 4,195 & 126 \\
\bottomrule
\end{tabular}
\caption{Annotation and query composition across the four ten-task LIBERO
suites. One query is created at each phase start. ``Invalid'' counts stored
\texttt{phase\_valid=false} labels.}
\label{tab:suite-stats}
\end{table}

\begin{table}[htbp]
\centering
\small
\begin{tabular}{lr}
\toprule
Statistic & Value \\
\midrule
Annotated phases / phase-start queries & 9,507 \\
Exact unique phase descriptions & 503 \\
Phases per episode, mean / median & 5.553 / 5 \\
Phases per episode, min / max & 2 / 11 \\
Valid / invalid flags & 9,381 / 126 \\
Duration in frames, mean / median & 29.21 / 23 \\
Duration in frames, 10th / 90th percentile & 3 / 59 \\
Queries with action padding & 2,714 \\
\bottomrule
\end{tabular}
\caption{Aggregate annotation statistics from the 9,507-row phase-start
manifest.}
\label{tab:annotation-stats}
\end{table}

\section{Chunk-Aligned Query Construction}

\subsection{Occupancy Labels}

For a query at frame $t$, let $\mathcal{C}_t$ contain the valid, non-padded action
indices in the next $H=32$ demonstration steps and let $g_\tau$ denote the
phase containing step $\tau$.  The occupancy of phase $k$ is
\[
 w_{t,k}=
 \frac{\left|\{\tau\in\mathcal{C}_t:g_\tau=k\}\right|}{|\mathcal{C}_t|}.
\]
All intersecting phases are retained, and occupancies are renormalized over
the valid steps.  Consequently, a short final chunk still sums to one and does
not assign semantic mass to padded actions.  Of the 9,507 phase-start queries,
2,714 contain action padding. A valid final chunk retains unit sample weight;
its shorter length changes the occupancy denominator, not its loss weight.

Each phase contributes one query at its start frame. A query is \emph{Pure}
when one stage occupies all valid future steps and \emph{Mixed} when two or
more stages overlap the chunk. Mixed-2 contains exactly two stages; Mixed-3+
contains at least three. These categories use the full valid action horizon,
not only the stage at the first step.

\subsection{Fixed Four-Token Teacher}

Each free-form description is encoded by the frozen UMT5 encoder
\citep{DBLP:conf/iclr/ChungGRTFNC23}. Its valid
token sequence is partitioned into four contiguous, non-empty intervals and
mean-pooled within each interval.  We center the pooled training-description
vectors and construct a fixed random projection from a seed-0 Gaussian matrix
followed by reduced QR decomposition; its 128 orthonormal rows are independent
of the semantic training data.  The teacher target is the
occupancy-weighted mixture of the projected stage slots:
\[
 T_{t,n}=R\left(\sum_k w_{t,k}\bar{h}_{k,n}-\mu\right),
 \qquad n\in\{1,\ldots,4\}.
\]
Thus $T_t\in\mathbb{R}^{4\times128}$ preserves coarse token order while
matching the temporal support of the action chunk. A valid sample must supply
all four pooled teacher slots; missing or incomplete teacher representations
raise an error.

\section{Teacher Matching}

\subsection{Episode-Local Retrieval}

The analysis uses all 1,712 annotated LIBERO training demonstrations and the
10,000-step CASD generator checkpoint. It measures representation matching on
training episodes, separately from the rollout success rates in the main paper.

For an episode $e$ containing $N_e$ annotated phases, the gallery contains the
$N_e$ fixed teacher targets constructed at those phase starts.  The
CASD generator predicts $P_i$ from the current observation, robot state, and
task instruction.  Phase boundaries and future occupancies are used only after
prediction to construct the targets and gallery candidates.

We use the negative normalized Stage-A regression distance as the matching
score:
\[
 s(i,j) =
 -\frac{\lVert P_i-T_j\rVert_F^2}
 {4\cdot128\cdot\sigma_T^2},
\]
where $\sigma_T$ is the RMS teacher scale stored with the fixed teacher.  The
correct item is the teacher context at the same phase start.  If multiple
candidates tie for the maximum score, Recall@1 assigns $1/m$ credit when the
correct candidate is among the $m$ tied maxima and zero otherwise.

For random guessing, query $i$ has expected recall $1/N_e$.
Chance Recall in Table~\ref{tab:recovery} is the average over queries;
each episode contributes in proportion to its number of queries.

\begin{table}[htbp]
\centering
\small
\begin{tabular}{lrrrr}
\toprule
Subset & Queries & Share & Chance & Recall@1 \\
\midrule
Pure     & 5,312 & 55.87\% & 0.185 & 0.430 \\
Mixed    & 4,195 & 44.13\% & 0.174 & 0.765 \\
Mixed-2  & 3,643 & 38.32\% & 0.175 & 0.761 \\
Mixed-3+ &   552 &  5.81\% & 0.166 & 0.790 \\
\bottomrule
\end{tabular}
\caption{Episode-local teacher matching at annotated training phase
starts. Mixed-2 and Mixed-3+ partition Mixed. Values are averaged over queries,
with fractional credit for top-score ties.}
\label{tab:recovery}
\end{table}

The corresponding Recall@1-to-chance ratios are $2.3$, $4.4$, $4.4$, and
$4.8$ for Pure, Mixed, Mixed-2, and Mixed-3+, respectively, computed before
rounding. Mixed queries cross stage boundaries within the 32-step action
horizon. The ratios also depend on gallery composition and do not measure
the relative difficulty of the subsets.

\FloatBarrier

\section{Training Configuration}

For LIBERO, the CASD generator has hidden width 128. Its inputs comprise a
432-dimensional frozen-VAE feature of the current frame, the eight-dimensional
robot state, and frozen UMT5 task embeddings. The teacher has four tokens of
width 128 and uses occupancy over the valid part of an $H=32$ action chunk.

Stage A is configured for 10,000 optimization steps, a global batch size of
128, learning rate $10^{-4}$, weight decay $10^{-2}$, and a cosine schedule.
The direct-regression and same-task hardest-negative ranking losses each
have coefficient 1. The minimum eligible teacher distance is 0.05, and the
margin ratio is 0.5.  Only CASD generator parameters are optimized.  The
frozen VAE, UMT5 encoder, teacher transformation, and policy backbone receive
no Stage-A updates. Stage A samples frame anchors throughout the annotated
training demonstrations and applies the binary mask defined above. The
9,507-row phase-start manifest is used only for retrieval analysis; stored
annotation flags do not filter this manifest either.

At Stage B, the selected CASD generator checkpoint is loaded and frozen.
Policy training recomputes its prediction from the current observation
rather than reading teacher tokens or cached future-dependent occupancies.
This keeps the CASD-token source identical during policy training and
deployment. The direct-regression and ranking losses are disabled in Stage B.

\paragraph{Stage-B injection.}
Fast-WAM-Joint \citep{DBLP:journals/corr/abs-2603-16666} contains 30 aligned
video/action expert blocks of widths 3072/1024. Each expert independently
projects the four shared predicted tokens to width 1024 with an affine map
and GELU. The resulting context is supplied to an independent rank-192
Q/K/V/O cross-attention adapter in every video and action block. These
adapters use 24 heads of dimension 128. For expert
width $d_e$, Q maps $d_e\!\rightarrow192\!\rightarrow3072$, K/V map
$1024\!\rightarrow192\!\rightarrow3072$, and O maps
$3072\!\rightarrow192\!\rightarrow d_e$. Thus the attention width is
$24\times128=3072$ in both experts, even though the action residual stream
has width 1024. The last linear layer of O is zero-initialized so the
pretrained function is unchanged at initialization.

\section{Policy Metrics and Result Sources}

\subsection{Success Aggregation}

For a task with $n$ evaluation episodes and binary outcomes $y_j$, success is
$100\,\sum_j y_j/n$. LIBERO reports the mean task success within each ten-task
suite and the mean of the four suite scores. RoboTwin reports the mean over
50 tasks separately for Clean and Rand.; its Average gives the two conditions
equal weight. The MolmoSpaces comparison reports the four manipulation
categories Pick, Pick-and-Place, Open, and Close, with equal weight in the
displayed average.

LIBERO-Plus uses a different aggregation. Let $n_c$ and $s_c$ denote evaluated
episodes and successes in perturbation category $c$. Its pooled score is
\[
 S_{\mathrm{pooled}}=100\,\frac{\sum_{c=1}^{7}s_c}{\sum_{c=1}^{7}n_c}.
\]
Categories have different sizes, so this is not an unweighted mean of the
seven displayed percentages. Table~\ref{tab:pooled-counts} gives the totals
behind the CASD rows. The Joint run's suite-summary value of 72.36 uses a
different aggregation and is not the pooled value in the main table.

\begin{table}[htbp]
\centering
\small
\begin{tabular}{@{}lrrr@{}}
\toprule
Fast-WAM model & Episodes & Successes & Pooled (\%) \\
\midrule
Uncond+CASD & 10,030 & 5,737 & 57.2 \\
Joint+CASD & 10,030 & 7,238 & 72.2 \\
IDM+CASD & 10,030 & 7,417 & 73.9 \\
\bottomrule
\end{tabular}
\caption{LIBERO-Plus aggregation from the seven category counts.}
\label{tab:pooled-counts}
\end{table}

Policy scores in the main paper use one decimal place. Differences between our LIBERO
variants and their references use the unrounded run averages: IDM+CASD is
98.85 and Uncond+CASD is 94.45, giving changes of $+0.85$ and $-3.15$ percentage
points. The main text rounds these changes to $+0.9$ and $-3.2$ points.

\subsection{RoboTwin Per-Task Results}

Table~\ref{tab:robotwin-per-task} reports success rates for all 50 RoboTwin
tasks using the Uncond, IDM, and Joint Fast-WAM variants with CASD under
Clean and domain-randomized (Rand.) evaluation. Our policies execute the first
28 actions of each predicted chunk before replanning from a new observation
(\texttt{replan\_steps=28}). Each task has equal weight within a condition, and Clean and
Rand. have equal weight in the overall average. The overall scores are
90.92\%, 92.39\%, and 92.96\% for Uncond+CASD, IDM+CASD, and Joint+CASD,
respectively, matching the rounded entries in the main paper.
Unaugmented reference results come from the published sources described below.

\subsection{Published References}

The unaugmented Fast-WAM rows in Tables~\ref{tab:libero-main} and~\ref{tab:robotwin} follow the official
report \citep{DBLP:journals/corr/abs-2603-16666} for the same named variants and
benchmark settings. Table~\ref{tab:libero-plus} uses external Fast-WAM checkpoints evaluated by
RIFT (Appendix~D, Table~5, version~2) \citep{zhang2026rift}; other baselines
follow the LIBERO-Plus report \citep{DBLP:journals/corr/abs-2510-13626}.

Sources for the remaining LIBERO and RoboTwin models appear in the main
paper. MolmoSpaces references use the official September~2, 2026 leaderboard
snapshot \citep{molmospaces2026leaderboard} and were not re-evaluated here.

\section{Paired Rollout Analysis}

The Figure~\ref{fig:libero-paired-case} rollout outcomes, action traces, videos, and CASD generator
predictions come from zero-indexed trial~2 of zero-indexed LIBERO-Long task~3:
put the black bowl in the bottom drawer of the cabinet and close it.
Fast-WAM-Joint succeeds in 49/50 trials and Fast-WAM-Joint+CASD in 50/50;
trial~2 is the paired failure/success case. Their recorded initial observations
and all 22 common inference seeds match. The displayed transition is
r15$\rightarrow$r16, from Move/place toward Close.

\paragraph{Behavioral signals.}
CASD-token change is the symmetric nearest-neighbor cosine distance between
adjacent $4\times128$ token sets. Motion divergence is the root-mean-square
difference over the first six non-gripper action coordinates, aligned within
the same replan and restricted to executed actions. Gripper mismatch is the
fraction of aligned gripper-sign commands that disagree. Token change and
motion divergence are each divided by their maximum over the 22 common
replans; gripper mismatch remains on its native $[0,1]$ scale.

\paragraph{Semantic fit.}
For each task-specific stage description $u_k$, let $U_k$ be its fixed,
four-slot projected teacher prototype. For visualization, we fit a student
context $P_t$ by solving
\[
 \min_{\alpha\geq0,\,\mathbf{1}^{\top}\alpha=1}
 \left\|P_t-\sum_k\alpha_k U_k\right\|_F^2.
\]
The implementation uses deterministic projected gradient descent on the
simplex. Prototype weights are then summed into the displayed semantic
categories. These are reconstruction coefficients, not annotated occupancies
or calibrated probabilities. They are computed after the rollout and are
never supplied to the policy.

\paragraph{Frame selection.}
The eight frames show both policies before the boundary, inside the r16
divergence chunk, in the following replan, and at the CASD completion step.
The manifest records the exact frame indices and selection rationale.
The figure connects a semantic transition with behavior in one selected
rollout, rather than measuring a causal effect of changing an individual token.

\section{Implementation Scope and Reproducibility}

The implementation described here covers the Fast-WAM annotation,
teacher-construction, training, and evaluation pipeline. The separate
DreamZero integration is outside this implementation description.

The source archive accompanying this preprint contains the manuscript and
final figures. It does not include training or evaluation code, model weights,
generated phase annotations, normalization statistics, embedding caches,
per-query retrieval results, or full rollout traces. The archive is therefore
not a self-contained reproduction of the reported experiments.

\newpage

For a complete teacher-matching record, the analysis scripts export a row per
query with suite, task, episode, phase start, validity, padding, overlapping
stages, and occupancies, plus per-query tie credit. Counts and chance baselines
can be reconstructed from such a manifest; Recall@1 additionally requires the
corresponding predictions or checkpoint and teacher artifact. Checkpoint
reuse also requires the matching model configuration and normalization artifacts.

\paragraph{Annotation variability.}
The LIBERO launcher requests the \texttt{gpt-5.4} alias through an
OpenAI-compatible service. The saved annotations, rather than new service
responses, are the inputs to subsequent training. Action-based snapping
refines boundary timing but does not verify every free-form description.
Stored annotation scores are heuristic, and no inter-annotator agreement is
reported.

\clearpage

\begin{table*}[p]
\centering
\small
\setlength{\tabcolsep}{7pt}
\begin{tabular}{@{}lrrrrrr@{}}
\toprule
& \multicolumn{2}{c}{Uncond+CASD} & \multicolumn{2}{c}{IDM+CASD} & \multicolumn{2}{c}{Joint+CASD} \\
\cmidrule(lr){2-3}\cmidrule(lr){4-5}\cmidrule(l){6-7}
Task & Clean & Rand. & Clean & Rand. & Clean & Rand. \\
\midrule
\texttt{adjust\_bottle} & 100 & 100 & 93 & 100 & 99 & 99 \\
\texttt{beat\_block\_hammer} & 97 & 98 & 100 & 96 & 99 & 95 \\
\texttt{blocks\_ranking\_rgb} & 100 & 98 & 100 & 99 & 100 & 99 \\
\texttt{blocks\_ranking\_size} & 88 & 88 & 83 & 84 & 78 & 92 \\
\texttt{click\_alarmclock} & 100 & 100 & 100 & 100 & 100 & 100 \\
\texttt{click\_bell} & 100 & 100 & 100 & 100 & 100 & 100 \\
\texttt{dump\_bin\_bigbin} & 96 & 100 & 98 & 98 & 95 & 95 \\
\texttt{grab\_roller} & 100 & 100 & 100 & 100 & 100 & 100 \\
\texttt{handover\_block} & 92 & 85 & 98 & 92 & 99 & 91 \\
\texttt{handover\_mic} & 99 & 99 & 98 & 99 & 100 & 100 \\
\texttt{hanging\_mug} & 38 & 49 & 49 & 54 & 42 & 49 \\
\texttt{lift\_pot} & 100 & 100 & 100 & 100 & 100 & 100 \\
\texttt{move\_can\_pot} & 84 & 89 & 100 & 98 & 97 & 94 \\
\texttt{move\_pillbottle\_pad} & 98 & 99 & 99 & 99 & 100 & 100 \\
\texttt{move\_playingcard\_away} & 100 & 100 & 100 & 99 & 100 & 100 \\
\texttt{move\_stapler\_pad} & 72 & 69 & 77 & 82 & 83 & 78 \\
\texttt{open\_laptop} & 98 & 100 & 98 & 98 & 95 & 99 \\
\texttt{open\_microwave} & 58 & 70 & 70 & 72 & 85 & 93 \\
\texttt{pick\_diverse\_bottles} & 83 & 85 & 89 & 87 & 90 & 92 \\
\texttt{pick\_dual\_bottles} & 93 & 91 & 100 & 98 & 100 & 99 \\
\texttt{place\_a2b\_left} & 92 & 95 & 93 & 97 & 93 & 98 \\
\texttt{place\_a2b\_right} & 96 & 95 & 96 & 96 & 97 & 93 \\
\texttt{place\_bread\_basket} & 92 & 94 & 90 & 98 & 91 & 97 \\
\texttt{place\_bread\_skillet} & 89 & 96 & 93 & 87 & 87 & 92 \\
\texttt{place\_burger\_fries} & 95 & 99 & 99 & 99 & 100 & 98 \\
\texttt{place\_can\_basket} & 72 & 72 & 75 & 67 & 71 & 72 \\
\texttt{place\_cans\_plasticbox} & 97 & 96 & 98 & 98 & 98 & 99 \\
\texttt{place\_container\_plate} & 96 & 99 & 98 & 98 & 98 & 100 \\
\texttt{place\_dual\_shoes} & 96 & 91 & 88 & 85 & 88 & 92 \\
\texttt{place\_empty\_cup} & 98 & 100 & 100 & 100 & 100 & 100 \\
\texttt{place\_fan} & 99 & 98 & 99 & 95 & 98 & 95 \\
\texttt{place\_mouse\_pad} & 82 & 82 & 95 & 94 & 93 & 92 \\
\texttt{place\_object\_basket} & 90 & 87 & 86 & 78 & 85 & 78 \\
\texttt{place\_object\_scale} & 96 & 94 & 90 & 95 & 97 & 94 \\
\texttt{place\_object\_stand} & 94 & 91 & 98 & 97 & 95 & 95 \\
\texttt{place\_phone\_stand} & 95 & 96 & 99 & 96 & 99 & 96 \\
\texttt{place\_shoe} & 95 & 99 & 94 & 97 & 96 & 95 \\
\texttt{press\_stapler} & 95 & 98 & 89 & 89 & 89 & 91 \\
\texttt{put\_bottles\_dustbin} & 90 & 95 & 93 & 97 & 92 & 96 \\
\texttt{put\_object\_cabinet} & 92 & 88 & 95 & 87 & 84 & 85 \\
\texttt{rotate\_qrcode} & 87 & 87 & 88 & 75 & 87 & 88 \\
\texttt{scan\_object} & 88 & 88 & 95 & 96 & 91 & 96 \\
\texttt{shake\_bottle} & 100 & 100 & 100 & 100 & 100 & 100 \\
\texttt{shake\_bottle\_horizontally} & 100 & 100 & 100 & 100 & 100 & 99 \\
\texttt{stack\_blocks\_three} & 91 & 94 & 98 & 96 & 98 & 94 \\
\texttt{stack\_blocks\_two} & 100 & 99 & 100 & 100 & 100 & 100 \\
\texttt{stack\_bowls\_three} & 82 & 78 & 75 & 85 & 80 & 83 \\
\texttt{stack\_bowls\_two} & 94 & 95 & 89 & 95 & 92 & 95 \\
\texttt{stamp\_seal} & 83 & 81 & 86 & 89 & 95 & 91 \\
\texttt{turn\_switch} & 52 & 61 & 77 & 70 & 77 & 84 \\
\midrule
Mean (50 tasks) & 90.48 & 91.36 & 92.56 & 92.22 & 92.66 & 93.26 \\
Overall (Clean/Rand.) & \multicolumn{2}{c}{90.92} & \multicolumn{2}{c}{92.39} & \multicolumn{2}{c}{92.96} \\
\bottomrule
\end{tabular}
\caption{RoboTwin 2.0 per-task success rates (\%) for the three CASD-augmented
Fast-WAM variants. Clean and Rand. correspond to the two evaluation conditions
in the main paper. Mean values are recomputed from all 50 task rows; Overall
averages the two condition means.}
\label{tab:robotwin-per-task}
\end{table*}

\end{document}